%% file: corl_2026_template_submission/example.tex
\documentclass{article}
\usepackage{graphicx}
\usepackage{wrapfig}
\usepackage[utf8]{inputenc}
\usepackage{enumitem}
\usepackage{amsmath} 
\usepackage{siunitx}

\PassOptionsToPackage{table}{xcolor}

\usepackage{xcolor}

\definecolor{goodfill}{HTML}{E1F5EE}
\definecolor{goodtext}{HTML}{04342C}
\definecolor{subfill}{HTML}{FAEEDA}
\definecolor{subtext}{HTML}{412402}

\usepackage[preprint]{corl_2026} 

\usepackage{booktabs}
\usepackage{amssymb}

\hypersetup{citecolor=[HTML]{6e1fab}}

\title{CHORUS: Decentralized Multi-Embodiment Collaboration with One VLA Policy}

\author{
  Ria Doshi \quad
  Tian Gao \quad
  Annie Chen \quad
  Chelsea Finn \quad
  Jeannette Bohg \\ \\
  Stanford University \quad \\ \\
  \textbf{\href{https://chorus-model.github.io}{\textcolor[HTML]{8f2cdb}{https://chorus-model.github.io}}}
}

\begin{document}
\maketitle

\vspace{-1em}

\begin{abstract}
\input{sections/abstract}
\end{abstract}


\begin{figure}[h]
  \centering
  \includegraphics[width=1\textwidth]{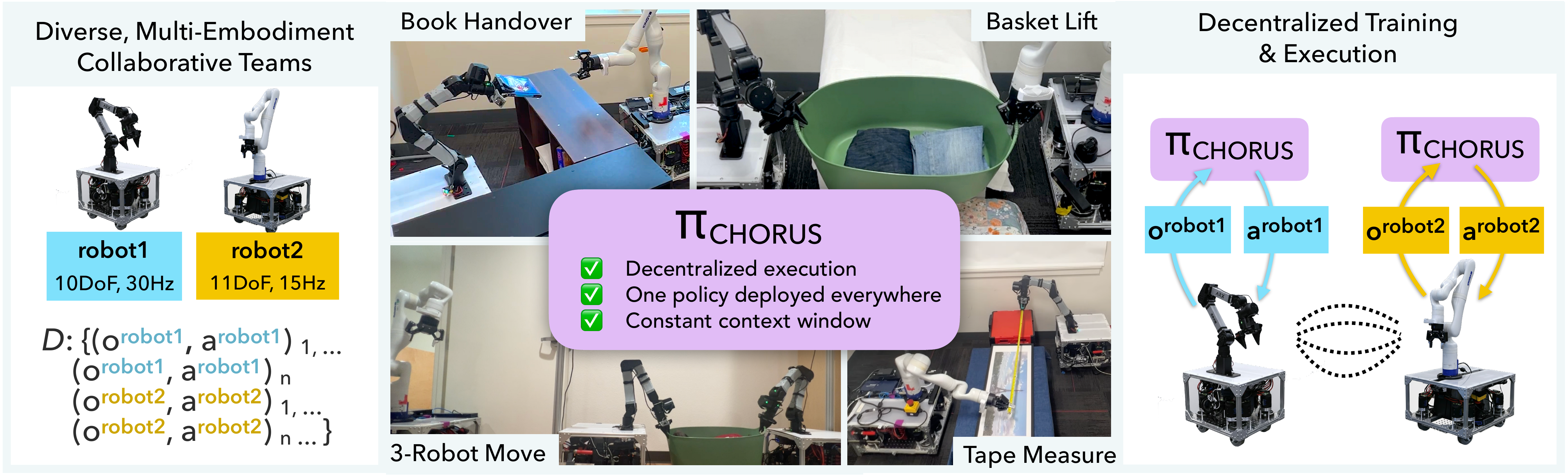}
  \caption{\footnotesize{We introduce \textbf{CHORUS}, a single VLA policy trained for decentralized, multi-embodiment collaboration. At inference, each robot runs a local copy of CHORUS, conditioned only on its \textit{own} observations and a robot-identifying prompt, enabling efficient and reactive collaboration without any inter-robot }communication.  }
  \vspace{1em}
  \label{fig:teaser}
\end{figure}



\section{Introduction}
\input{sections/introduction}

\section{Related Work}
\input{sections/related-works}

\section{Designing a VLA Policy for Decentralized Multi-Robot Control}
\input{sections/method.tex}

\section{Experiments}
\label{sec:experiments}
\input{sections/experiments.tex}

\section {Discussion and Conclusion}
\input{sections/conclusion.tex}



\clearpage
\acknowledgments{This work is supported by the Stanford Institute for Human-Centered Artificial
Intelligence (HAI), the National
Science Foundation (NSF) under Grant Numbers 2327974 and 2342246, and ONR grant N00014-22-1-2621, with fellowship support from the Department of Defense (DoD) through the National Defense Science \& Engineering Graduate (NDSEG) Fellowship program and the Stanford Graduate Fellowship program. 

We thank Cherie Ho and Jimmy Wu for assistance with robot setup; Jason Choi for discussions in the early stages of the project; and Alberta Longhini, Priya Sundaresan, William Chen, and Satvik Sharma for feedback on earlier drafts of the paper. 
 }


\bibliography{chorus-new}  

\newpage
\section{Appendix}
\input{sections/appendix}

\end{document}

%% file: sections/abstract.tex
Multi-robot collaboration allows robots to efficiently take on a wide range of tasks, from moving a couch through a doorway to assembling structures on a construction site. However, achieving such coordination in mobile multi-robot settings remains challenging: centralized methods conditioned on the combined observations of a team scale poorly with team size, and decentralized methods that train one policy per robot often require explicit alignment procedures or information sharing at inference time to overcome partial observability. Our key insight is that the visuomotor priors of pretrained vision-language-action (VLA) models should enable reactive, decentralized collaboration from each robot's local observations alone, without inference-time assumptions. We propose CHORUS, a framework that adapts a single VLA backbone to control diverse, multi-robot teams. At inference time, each robot runs an independent copy of CHORUS, conditioned only on its own observations and a robot-identifying prompt. In real-world experiments including mobile tape measurement, library book handovers, and laundry basket lifting, CHORUS achieves a 64\% point improvement over decentralized, from-scratch models, improves reactivity to teammate behavior by 40\% points, and outperforms centralized baselines. Together, these results show that a shared VLA backbone is capable of achieving decentralized multi-robot collaboration, without per-robot policies or inter-robot communication at inference.


%% file: sections/introduction.tex
Collaboration is a key feature of human intelligence. To collaboratively accomplish tasks, humans interpret body language, respond to visual cues, and adjust their actions on the fly. Robots that collaborate effectively must do the same, often as part of heterogeneous teams whose members are best suited for different roles and differ in kinematics. If we can imbue robots with such an ability to work with one another, we take a meaningful step towards true collaborative robot intelligence: multi-embodiment cleaning teams, construction crews, apartment movers, and much more.

Achieving this vision of collaborative robot intelligence requires a policy architecture capable of coordinating diverse robot morphologies together. One approach is to train a single centralized policy that conditions on team-wide observations and produces actions for all robots in a single forward pass \cite{tungLearningMultiArmManipulation2021, aljalboutCLASCoordinatingMultiRobot2023,zhaoLearningFineGrainedBimanual2023a}; however, the context window and action space grow with team size, and the policy requires inter-robot communication at inference to collate a team-wide observation. Decentralized policies offer a compelling alternative: by training a separate policy for each robot, context size stays constant as teams grow, robots can run independently at different control frequencies, and small temporal mismatches between teammates are tolerated \cite{amatoIntroductionCentralizedTraining2024a, dongMIMICDMultimodalImitation2025, heLatentTheoryMind2025a}. A key tradeoff of decentralization is partial observability: each robot must coordinate with teammates, but act only on local observations. Prior decentralized methods tackle this by introducing shared assumptions at execution time, such as conditioning on proprioceptive state from teammates \cite{dongMIMICDMultimodalImitation2025}, shared third-person cameras, or online alignment procedures \cite{heLatentTheoryMind2025a}; these approaches impose requirements at inference time that can be costly. Beyond execution mode, teammates may differ in sensors, control frequency, and kinematics, requiring an architecture that seamlessly operates across varying observation and action spaces.

Our key insight is that strong visuomotor priors may be sufficient to enable decentralized, multi-embodiment collaboration without alignment or communication at inference. Pretrained vision-language-action (VLA) models \cite{kimOpenVLAOpenSourceVisionLanguageAction2025, zitkovichRT2VisionLanguageActionModels2023}, trained on diverse robots and data, have demonstrated such priors, albeit in single-robot domains. Much of this data is bimanual, and bimanual manipulation can be viewed as a simplified form of multi-robot collaboration, though one in which the arms are physically coupled and controlled in a centralized manner. In this work, we investigate whether finetuning a pretrained VLA on multi-robot data yields fully decentralized collaboration among separate, mobile embodiments, and further, whether a single policy conditioned on a robot identity prompt eliminates the need to train per-robot policies entirely.

We present CHORUS, a method that finetunes a pretrained VLA for decentralized, multi-robot collaboration (Figure \ref{fig:teaser}). At inference, each robot runs an independent copy of CHORUS, conditioned only on its own observations along with a robot-specific identity prompt. \textit{No cameras, proprioceptive states, or communication channels are shared among robots}, making CHORUS more deployable than centralized alternatives and cheaper to train than per-robot decentralized policies.

\textbf{In summary, CHORUS provides a recipe for adapting a pretrained VLA backbone into a single policy for decentralized multi-embodiment collaboration.} In extensive, real-world evaluations, CHORUS outperforms the leading, from-scratch imitation learning approach \cite{chiDiffusionPolicyVisuomotor2023} by 64 percentage points in success rate, reacts to teammate behavior nearly 2x more effectively than a VLA backbone finetuned per-robot, and outperforms centralized policies that condition on the full observation space of the team. We further demonstrate that the same policy scales to larger, three-robot teams in the real world, achieving 90\% task success without any architectural change.

%% file: sections/related-works.tex
\paragraph{Classical collaborative manipulation.} Approaches to collaborative manipulation fall broadly into two categories. Centralized methods \citep{tungLearningMultiArmManipulation2021, aljalboutCLASCoordinatingMultiRobot2023, zhaoLearningFineGrainedBimanual2023a, fuMobileALOHALearning2024, yuBiKCKeyposeConditionedConsistency2024, xuBridgingDomainGap2023} condition on \textit{all} of the robots' observations jointly and produce actions for the whole team. Decentralized methods \citep{loweMultiAgentActorCriticMixed2017, yuSurprisingEffectivenessPPO2022b, amatoIntroductionCentralizedTraining2024a, heLatentTheoryMind2025a, dongMIMICDMultimodalImitation2025, vatnsdalScalableMultiAgent2025} condition on only on the robot's local observation and produce an action for just that robot. CHORUS targets decentralized execution because it requires no inter-robot communication at runtime and scales naturally to larger teams without inflating the observation and action spaces. One body of work addresses decentralized, collaborative manipulation through control and planning \citep{khatibCoordinationDecentralizedCooperation1996, sugarDecentralizedControlCooperating1998, changAugmentedObjectModel2000, wangObjectClosureManipulation2002, finkCompositionVectorFields2007, finkMultirobotManipulationCaging2008, wangKinematicMultirobotManipulation2016, culbertsonDecentralizedAdaptiveControl2018, tallamrajuMotionPlanningMultiMobileManipulator2019a, muvvalaStochasticGamesInteractive2024}. These methods deliver guarantees on many cooperative tasks with known dynamics \citep{tallamrajuMotionPlanningMultiMobileManipulator2019a, mellingerCooperativeGraspingTransport2013, tagliabueCollaborativeTransportationUsing2017}, but often assume access to object models and contact geometry; scaling them to diverse environments and behaviors remains an open problem.

\paragraph{Learning-based single-robot manipulation.} In contrast to classical approaches, vision-language-action (VLA) models have shown that a single policy can generalize across many tasks, scenes, and embodiments when trained on sufficiently diverse data \citep{zitkovichRT2VisionLanguageActionModels2023, kimOpenVLAOpenSourceVisionLanguageAction2025, driessPaLMEEmbodiedMultimodal2023a, liCogACTFoundationalVisionLanguageAction2024a, wenDexVLAVisionLanguageModel2025a, szotMultimodalLLMsGeneralist2025, zawalskiRoboticControlEmbodied2025a}, supported by cross-embodiment datasets \citep{oneillOpenXEmbodimentRobotic2024, ghoshOctoOpenSourceGeneralist2024, doshiScalingCrossEmbodiedLearning2025, khazatskyDROIDLargeScaleInTheWild2024, walkeBridgeDataV2Dataset2023, dasariRoboNetLargeScaleMultiRobot2020a, fangRH20TComprehensiveRobotic2024} and expressive action-decoding methods  \citep{chiDiffusionPolicyVisuomotor2023, zhaoLearningFineGrainedBimanual2023a}. These results suggest a path toward a foundation model for robot manipulation, but the focus has remained on single-robot tasks. We investigate to what extent these models can be used for decentralized, multi-robot collaboration, \textbf{\textit{a setting entirely out of distribution from pretraining}.}

\paragraph{Learning-based multi-robot manipulation.} Both reinforcement learning (RL) \citep{mnihHumanlevelControlDeep2015, schulmanTrustRegionPolicy2015, schulmanProximalPolicyOptimization2017, haarnojaSoftActorCriticOffPolicy2018} and imitation learning \citep{pomerleauALVINNAutonomousLand1988, schaalLearningDemonstration1996, argallSurveyRobotLearning2009, rossReductionImitationLearning2011, hoGenerativeAdversarialImitation2016} have been explored in multi-agent literature. Multi-agent RL (MARL) methods often share critics or mix networks during training while executing on local observations \citep{loweMultiAgentActorCriticMixed2017, sunehagValueDecompositionNetworksCooperative2018, wangQPLEXDuplexDueling2020, yuSurprisingEffectivenessPPO2022b, amatoIntroductionCentralizedTraining2024a, tampuuMultiagentCooperationCompetition2017}, but they operate in low-dimensional state spaces and rely on simulation for the sample budget MARL requires, complicating transfer to vision-based robots in the real world. Another body of works applies imitation learning directly: MIMIC-D \citep{dongMIMICDMultimodalImitation2025} learns diffusion policies \citep{chiDiffusionPolicyVisuomotor2023} for collaborative tasks, but conditions on the state of all robots in the team; LatentToM \citep{heLatentTheoryMind2025a} similarly trains diffusion policies and uses a latent theory-of-mind module involving an online alignment procedure and shared third-person camera view. \textbf{CHORUS instead presents a recipe for adapting a\textit{ pretrained VLA backbone} for multi-robot collaboration, without requiring any shared camera view, team proprioceptive state, or online alignment procedure.}


\paragraph{Task decomposition and role assignment.} A complementary line of work uses large language models to decompose collaborative tasks into subtasks and assign roles to team members \citep{mandiRoCoDialecticMultiRobot2024, chenScalableMultiRobotCollaboration2024a, ichterCanNotSay2023, huangInnerMonologueEmbodied2023, liangCodePoliciesLanguage2023a}. These systems address the higher-level question of which robot does what task, treating low-level execution as a given. Thus, this line of work is complementary to ours: CHORUS takes such a task specification as input and subsequently produces low-level actions for each robot.

%% file: sections/method.tex
\begin{figure}[t]
  \centering
  \includegraphics[width=\textwidth]{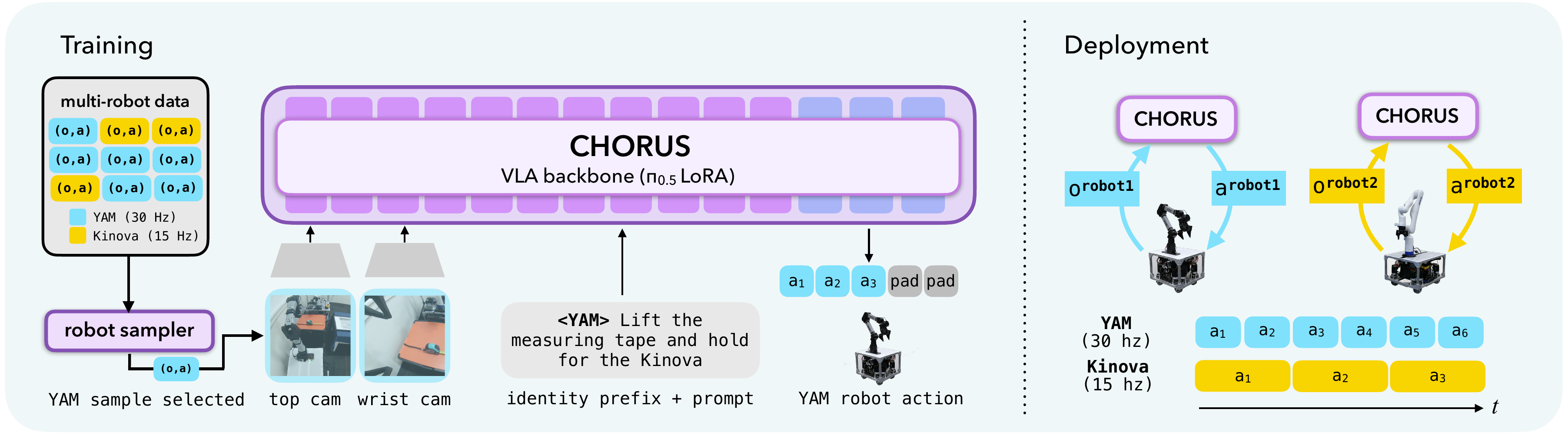}
  \caption{\textbf{CHORUS overview.} \emph{Training:} a single $\pi_{0.5}$ VLA is finetuned using LoRA on multi-robot data. The robot sampler draws one robot's $(o, a)$ per step. The policy conditions on this robot's identity prompt and predicts a padded action to accommodate different embodiments. \emph{Deployment:} the shared weights run independently on each robot, yielding fully decentralized execution at inference.}
  \label{fig:architecture}
\end{figure}
We introduce CHORUS, a single VLA policy that controls each robot in a team from its own observations, assuming no inter-robot communication at inference, and scales to larger, multi-embodiment teams. Designing such a policy requires three decisions: how to collect and structure multi-robot demonstrations, how to train a single policy for a diverse team, and how to deploy it at inference (Figure \ref{fig:architecture}). We address each in turn, building on the $\pi_{0.5}$ pretrained VLA backbone~\cite{intelligencep_05VisionLanguageActionModel2025}. 

\subsection{Multi-robot demonstrations and task design}

\textbf{Data collection.}  We collect training data on single-robot mobile manipulators via the TidyBot++ teleoperation interface~\cite{wuTidyBotOpenSourceHolonomic2025}. For two-robot teams, a single human operator controls each robot with a phone in each hand; for three-robot teams, a second human operator controls the third robot.

A multi-robot episode is a synchronously logged trajectory
\begin{equation}
    \xi = \left\{ \left(o_r^t,\, a_r^t \right) \right\}_{r \in [N],\, t \in [T]},
\end{equation}
where $N$ is the number of robots, $T$ the episode length (which may vary in different trajectories), $o_r^t$ robot $r$'s observation at time $t$, and $a_r^t$ the corresponding commanded action.

From each episode, we extract per-robot training tuples $(o_r^t, A_r^t, c_r)$, where $A_r^t = (a_r^t, \ldots, a_r^{t+H-1})$ is robot $r$'s action chunk of horizon $H$ and $c_r$ is the robot-identifying prompt for robot $r$ (Section~\ref{sec:policy}). All robots' tuples across all episodes are pooled into a single training set $\mathcal{D}$. \textbf{\textit{The model is never given a joint observation across robots:}} at training time, each tuple carries only one robot's local observation, and at inference, each robot independently queries the policy on its own.


\textbf{Collaboration strategy.} Because each robot conditions only on its own observation, $o_r$ must contain sufficient visual information to predict $A_r$. In many cases, we find that a collaborative task can be completed in multiple ways; we choose strategies to ensure that it is possible to complete the task in a decentralized manner. On each robot, we collect data so that the relevant teammate and workspace are visible from at least one of the robot's views throughout the task. Choosing such collaboration strategies during data collection makes decentralized execution feasible for the tasks we study.



\subsection{Policy design and training}
\label{sec:policy}

We train a single policy $\pi_\theta(A_r \mid o_r, c_r)$ shared across all robots in the team, as a shared policy is more efficient to train than $N$ embodiment-specific policies (Table \ref{tab:baseline_breakdown}). Beyond efficiency, we show that this shared backbone confers an additional performance advantage in in Section \ref{weightshare}.

\textbf{Cross-embodiment input format.} We inherit the backbone's cross-embodiment format: padded action vectors of dimension 32 and a variable number of image tokens per observation~\cite{intelligencep_05VisionLanguageActionModel2025}. This allows the policy to handle varying observation and action dimensions without architectural change.

\textbf{Robot-identifying prompt.} For a single shared policy to control every robot in the team, it must be able to identify which robot it is currently controlling. We supply this information through a robot-identifying prompt $c_r$ prepended to the model input at every timestep (see Figure \ref{fig:architecture}). The prompt names the embodiment (\texttt{<ARX>}, \texttt{<Kinova>}, etc.), specifies its role, and allows us to condition the forward pass on a known robot identity rather than forcing the policy to infer it from the observation. 

\textbf{Robot Sampler.} The robot sampler composes each training batch from single-robot tuples $(o_r^t, A_r^t, c_r)$ drawn independently from $\mathcal{D}$, with sampling weights governing how frequently each embodiment is drawn. In the two-robot setting, the embodiments are teleoperated at control frequencies differing by at most a factor of two, so we sample uniformly. In the three-robot case, the two YAM arms together generate actions at roughly four times the Kinova's rate, so uniform sampling would leave the Kinova undertrained; thus, we upweight Kinova tuples by a factor of two.

\textbf{Training objective and optimization.} We optimize the flow-matching loss inherited from the backbone~\cite{intelligencep_05VisionLanguageActionModel2025} over the pooled single-robot dataset $\mathcal{D}$:
\begin{equation}
    \mathcal{L}(\theta) = \mathbb{E}_{(o_r, A_r, c_r) \sim \mathcal{D},\, \tau \sim U(0,1),\, \epsilon \sim \mathcal{N}(0, I)} \left[ \left\| v_\theta(A_r^\tau,\, \tau \mid o_r, c_r) - (\epsilon - A_r) \right\|_2^2 \right],
\end{equation}
where $A_r^\tau = \tau A_r + (1-\tau)\epsilon$ is the noised action chunk and $v_\theta$ is the predicted velocity. The expectation is taken over single-robot tuples: the loss never sees a joint $(o, A)$ across robots.

We fine-tune the backbone with LoRA adapters~\cite{huLoRALowRankAdaptation2021a} of rank 16 and 32 applied to the VLM and the action expert respectively. We optimize with AdamW~\cite{loshchilovDecoupledWeightDecay2019a} under a cosine learning rate schedule. Additional training details are provided in Appendix A.

\subsection{Decentralized deployment}
At inference, each robot runs an independent instance of CHORUS on its own local observations. Concretely, at each timestep $t$, robot $r$ samples
\begin{equation} \label{eq:dec}
    A_r^t \sim \pi_\theta(\,\cdot \mid o_r^t,\, c_r)
\end{equation}
using only its own observation $o_r^t$ and identity prompt $c_r$. No information is exchanged between robots at runtime; coordination must emerge through each robot's visual perception of its teammates. In contrast, centralized formulations model the joint distribution $\pi(A_1^t, \ldots, A_N^t \mid o_1^t, \ldots, o_N^t)$, whose input and output dimensionalities grow with $N$.

Local conditioning in Eq.~\ref{eq:dec} supports \textbf{asynchronous execution}, which we use in our evaluations: each robot advances through its action chunks independently, so the Kinova may be executing an action from chunk $A_\text{Kinova}^{t}$ while the ARX executes one from $A_\text{ARX}^{t'}$, with $t \neq t'$. This matters in real-world deployments, where robots often sit on separate networks or compute instances and experience uneven round-trip latencies. While a centralized policy must wait for the slowest link at every joint query, CHORUS lets each robot progress as fast as its own connection allows, so long as the discrepancies stay small: large latency gaps will still desynchronize the robots and drift their behavior from the data distribution, but decentralized execution absorbs the minor gaps gracefully. Furthermore, our decentralized framework promotes flexible hosting: each robot can host its own copy of CHORUS, or queries from multiple robots can be batched on a shared compute instance.

For teams with varying control rates, we scale each robot's chunk size proportionally to its control rate; this ensures all robots plan over the same time horizon (Figure~\ref{fig:architecture}). For example, since the YAM runs at double the Kinova's rate (30Hz vs.\ 15Hz), we double the YAM's chunk size at execution.

%% file: sections/experiments.tex
\begin{figure}[t]
  \centering
  \vspace{-2em}
  \includegraphics[width=\textwidth]{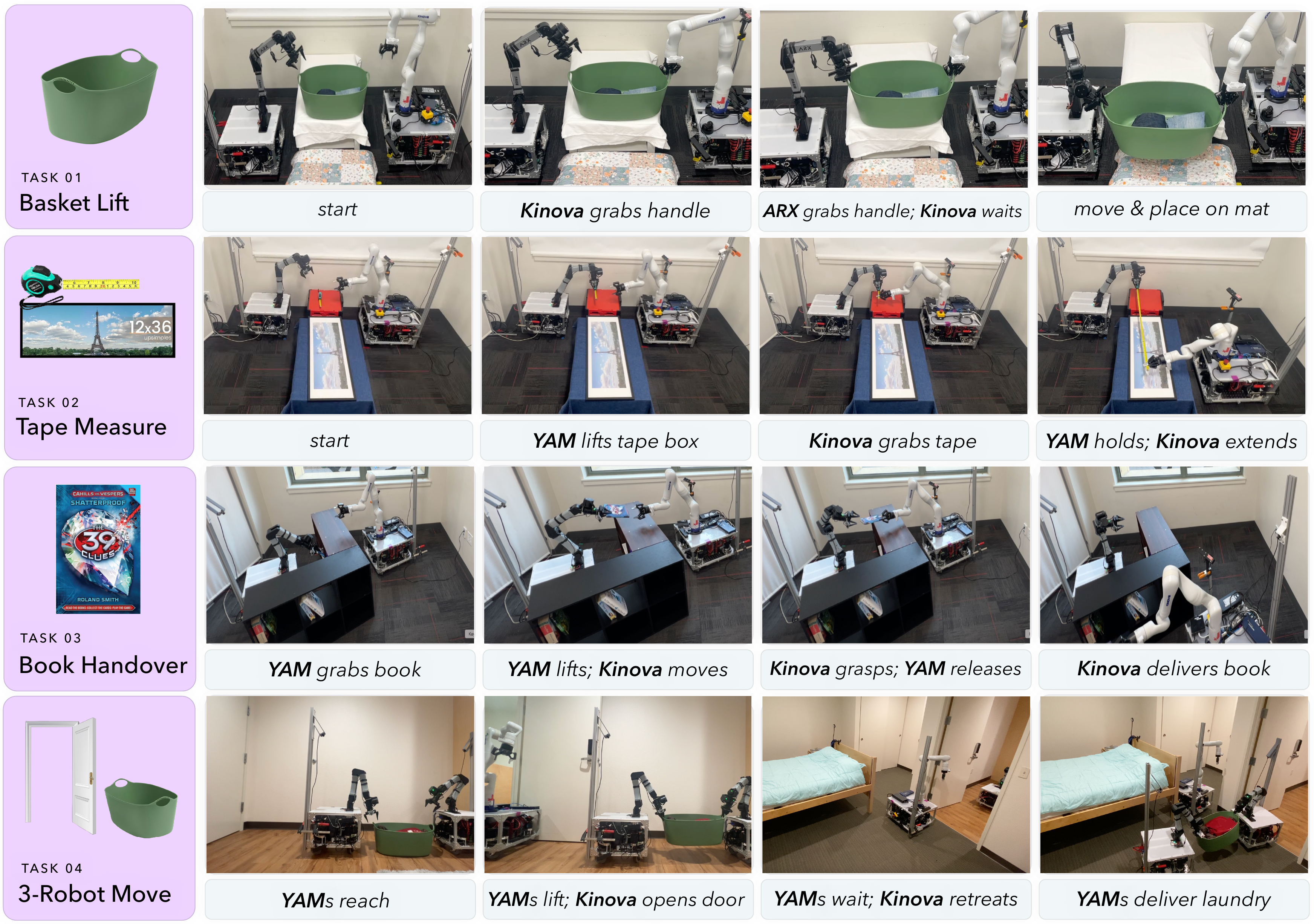}
  \caption{\textbf{Evaluation tasks.} We evaluate on a suite of multi-embodiment collaboration tasks: basket lifting, tape measuring, book handover, and 3-robot move. Note that the captions  describe task progression and are \textit{not} subtask prompts; we condition on one prompt per robot for the entire task.}
  \label{fig:tasks}
\end{figure}

Our experiments test whether CHORUS, a single VLA policy conditioned on each robot’s local observations, enables decentralized, multi-embodiment collaboration. We address four questions:
\begin{enumerate}[leftmargin=*, itemsep=0em]
    \itemsep0em
    \item Does a pretrained backbone provide a meaningful advantage over the leading from-scratch imitation learning method on multi-robot collaboration?
    \item Does sharing a set of weights across robots confer better reactivity to teammate behavior? 
    \item How does CHORUS compare to a centralized policy that conditions on the observations of the entire team jointly?
    \item How well does CHORUS scale to larger, three-robot teams?
\end{enumerate}

\begin{table}[t]
  \centering
  \begin{tabular}{l cc cc cc}
    \toprule
    & \multicolumn{2}{c}{VLA Centralized}
    & \multicolumn{2}{c}{CHORUS (w/o WS)}
    & \multicolumn{2}{c}{\textbf{CHORUS}} \\
    \cmidrule(lr){2-3} \cmidrule(lr){4-5} \cmidrule(lr){6-7}
    & $N\!=\!2$ & $N\!=\!3$ & $N\!=\!2$ & $N\!=\!3$ & $N\!=\!2$ & $N\!=\!3$ \\
    \midrule
    Total parameters $(\downarrow)$
      & \cellcolor{goodfill}\textcolor{goodtext}{3\,B} & \cellcolor{goodfill}\textcolor{goodtext}{3\,B}
      & \cellcolor{subfill}\textcolor{subtext}{6\,B} & \cellcolor{subfill}\textcolor{subtext}{9\,B}
      & \cellcolor{goodfill}\textcolor{goodtext}{\textbf{3\,B}} & \cellcolor{goodfill}\textcolor{goodtext}{\textbf{3\,B}} \\
    Context window $(\downarrow)$
      & \cellcolor{subfill}\textcolor{subtext}{$2C$} & \cellcolor{subfill}\textcolor{subtext}{$3C$}
      & \cellcolor{goodfill}\textcolor{goodtext}{$C$} & \cellcolor{goodfill}\textcolor{goodtext}{$C$}
      & \cellcolor{goodfill}\textcolor{goodtext}{$\boldsymbol{C}$} & \cellcolor{goodfill}\textcolor{goodtext}{$\boldsymbol{C}$} \\
    \bottomrule
  \end{tabular}
  \vspace{6pt}
  \caption{\textbf{Baseline Breakdown.} For a team of $N$ robots, where each robot contributes a context window of size $C$ (i.e. camera views + proprioception), CHORUS keeps parameters \& context window length constant in team size. In contrast, VLA 
  Centralized scales linearly in context length, while CHORUS (w/o Weight Sharing) scales linearly in total parameters.}
  \vspace{-1em}
  \label{tab:baseline_breakdown}
\end{table}

\paragraph{Baseline design.} To our knowledge, no prior work applies a VLA backbone to multi-robot collaboration. We therefore construct our baselines around CHORUS's two main design choices: decentralized inference and weight-sharing across robots  (Table~\ref{tab:baseline_breakdown}). \emph{VLA Centralized} conditions a single policy on the entire team's observations, removing decentralization while keeping weights shared, but requiring inter-robot communication at inference to collate the team-wide observation. This tests whether centralization, with strictly more information, yields a performance advantage in practice (Q3). \emph{CHORUS (w/o WS)} ablates weight sharing (WS) by training a separate policy per robot while keeping inference decentralized. This isolates the contribution of the shared backbone (Q2). Note that, regardless of task performance, both baselines scale poorly with team size --- \textit{VLA Centralized} in context length, and \textit{CHORUS (w/o WS)} in total parameters --- while CHORUS remains constant (Table~\ref{tab:baseline_breakdown}). We also compare against decentralized diffusion~\cite{chiDiffusionPolicyVisuomotor2023}, a standard from-scratch imitation learning recipe, to test whether VLA pretraining is necessary at all (Q1).

\paragraph{Setup.} In our experiments, we use several TidyBot \cite{wuTidyBotOpenSourceHolonomic2025} mobile manipulators: the Kinova (11\,DoF), ARX (10\,DoF), and YAM (10\,DoF). Our tasks include laundry basket lifting, mobile tape measurement, mobile book handover, and 3-robot transport (Figure~\ref{fig:tasks}). For each, we collect 25-45 demos and perform 10-18 rollouts, adding distractors to a subset of scenes at evaluation. Each robot receives one prompt for the entire task. Additional evaluation details are reported in Appendix B.

\begin{wrapfigure}{r}{0.55\textwidth}
  \centering
  \vspace{-2em}  \includegraphics[width=0.55\textwidth]{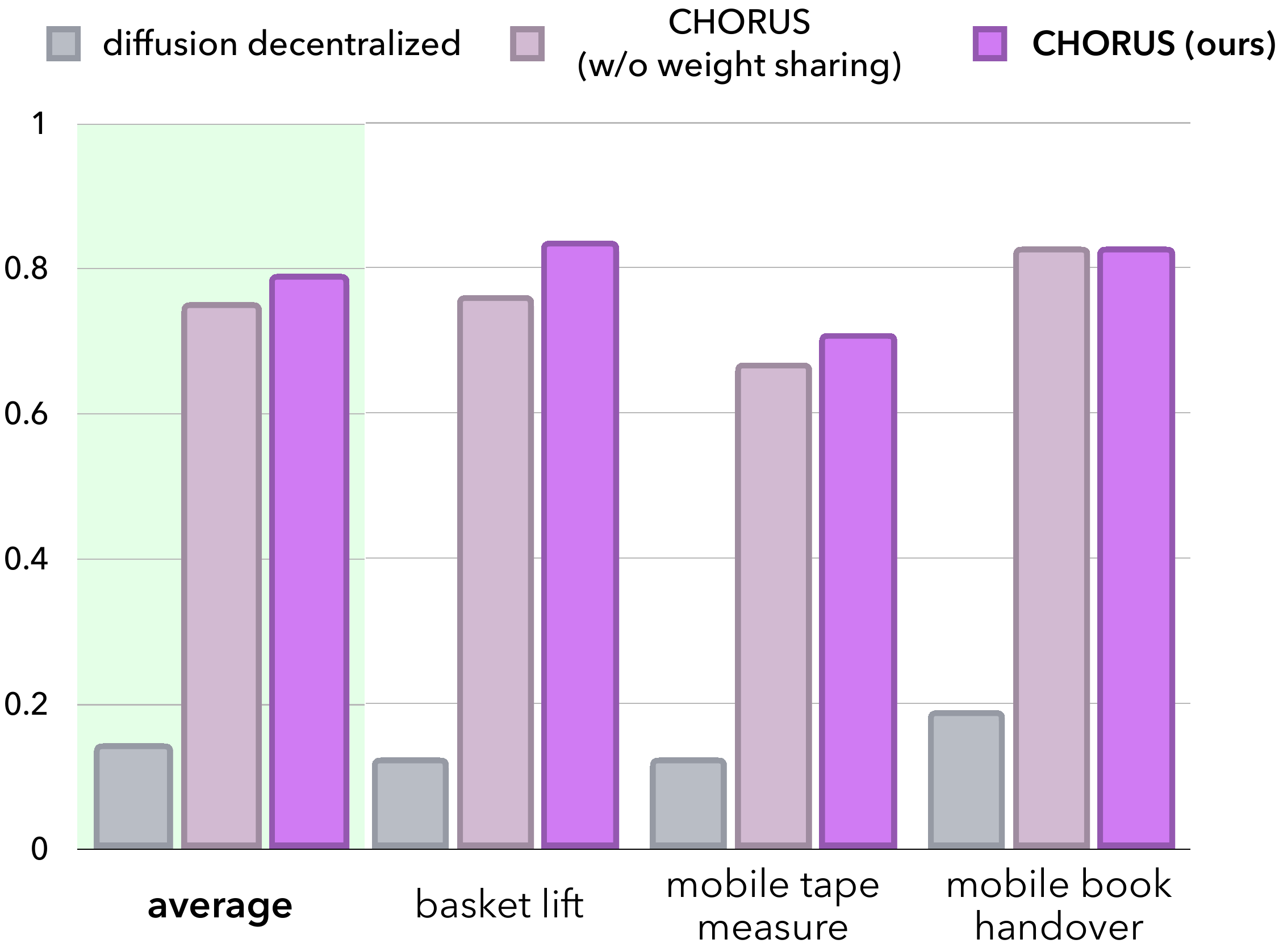}
  \caption{\textbf{Pretrained backbone comparison.} Both VLA-based methods significantly outperform decentralized diffusion, with CHORUS leading by 64 percentage points in mean success rate.}
  \vspace{-1em}
  \label{fig:decentralized-comparison}
\end{wrapfigure}

\subsection{Effect of the pretrained backbone (Q1)}

We first ask whether a pretrained backbone provides any meaningful advantage on multi-robot collaboration, given that this setting is OOD from pretraining. We compare CHORUS and CHORUS (w/o WS) against decentralized diffusion, which trains a separate diffusion policy per robot and represents the standard from-scratch imitation-learning recipe applied to the collaborative setting. 
Both VLA-based methods significantly outperform decentralized diffusion by a wide margin (Figure~\ref{fig:decentralized-comparison}), with \textbf{CHORUS leading by 64 percentage points in mean task success rate}. We observe two recurring failure modes. On tape measurement, distractor objects placed in the scene cause the diffusion policies to confuse the tape measure with non-target items. Across all three tasks, the diffusion policies exhibit a characteristic \emph{mismatch} pattern in which one robot proceeds with its half of the interaction before the other has caught up: on basket lifting, for example, the Kinova grabs its handle and begins to move before the ARX has grasped the other handle, causing the basket to slip. We take these results as evidence that a pretrained backbone supplies visuomotor priors that are indeed valuable for multi-robot collaboration, \textbf{\textit{even though collaborative tasks lie outside the pretraining distribution.}}

\subsection{Effect of weight-sharing on teammate reactivity (Q2)}
\label{weightshare}

\begin{figure}[t]
\centering
\begin{minipage}[c]{0.45\linewidth}
  \centering
  \includegraphics[width=\linewidth]{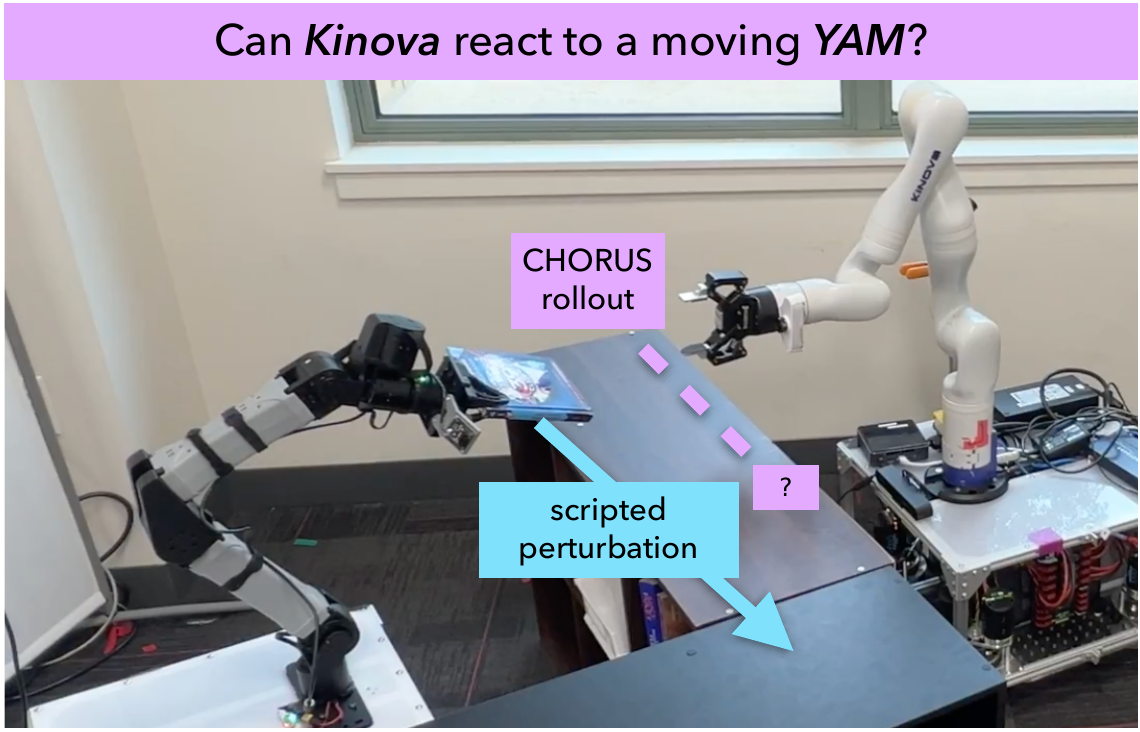}
\end{minipage}%
\hfill
\begin{minipage}[c]{0.5\linewidth}
  \centering
  \begin{tabular}{lcc}
  \toprule
   & CHORUS (w/o WS) & \textbf{CHORUS} \\
  \midrule
  Left perturb  & 3/10 & \textbf{8/10} \\
  Right perturb & 6/10 & \textbf{9/10} \\
  \midrule
  Total              & 9/20 & \textbf{17/20} \\
  \bottomrule
  \end{tabular}
\end{minipage}

\caption{\textbf{Assessing teammate reactivity.} The YAM (left) is perturbed laterally in a scripted trajectory; the Kinova (right) runs the policy and must adapt to the YAM's motion to complete the handover. Over 20 trials, CHORUS recovers 40\% more often. In settings where a teammate is perturbed, this result shows how weight-sharing can lead to better teammate reactivity.}
\label{fig:perturbation}

\end{figure}

Having established that a pretrained backbone confers a significant performance boost, the next question we ask is whether sharing such a VLA policy across robots, rather than training a separate copy per robot, yields any additional benefit. The \textit{CHORUS (w/o Weight-Sharing)} ablation isolates this question precisely: same pretrained initialization, but trained as two separate policies for each robot, instead of one. In settings that are primarily in-distribution from a motion standpoint, CHORUS and the per-robot framework perform comparably (Figure~\ref{fig:decentralized-comparison}), 

However, training on all robots' perspectives of every interaction should induce a shared representation that models the teammates' actions, whereas the per-robot policies, each trained only on its own robot's data, have no representational incentive to do so. We hypothesize that this is especially important in tasks where anticipating or reacting to teammates' behaviors is critical. To test this hypothesis, we design an experiment to assess reactivity on the mobile handover task (Figure~\ref{fig:perturbation}). Here, we script the YAM arm to grasp and lift the book, and laterally perturb it to the left or right; the Kinova runs the trained policy. To succeed, the Kinova must accurately track the YAM robot, matching its lateral movement, and complete the grasp.

Across twenty perturbation trials, we find that \textbf{CHORUS is nearly 2x more effective in successfully recovering the handover} (Figure~\ref{fig:perturbation}) over the CHORUS w/o WS policy. The main failure mode of the non-weight-shared policy is early grasping, wherein the Kinova grasps the book in the most in-distribution location, rather than tracking the new position of the YAM. We interpret this performance gap as evidence that using a single set of weights trained on both robots' perspectives can indeed improve reactivity to teammate behavior.

\subsection{Comparison to centralized policy framework (Q3)}

We now compare CHORUS to a centralized VLA formulation that conditions on concatenated observations from both robots and produces actions for both robots jointly. Since a centralized policy has access to strictly more information than any decentralized method, it should, in principle, serve as an \textbf{\textit{upper bound}} on collaborative performance. We now ask whether such a policy realizes this upper bound in practice.

\begin{wrapfigure}{r}{0.5\textwidth}
  \centering
\includegraphics[width=0.5\textwidth]{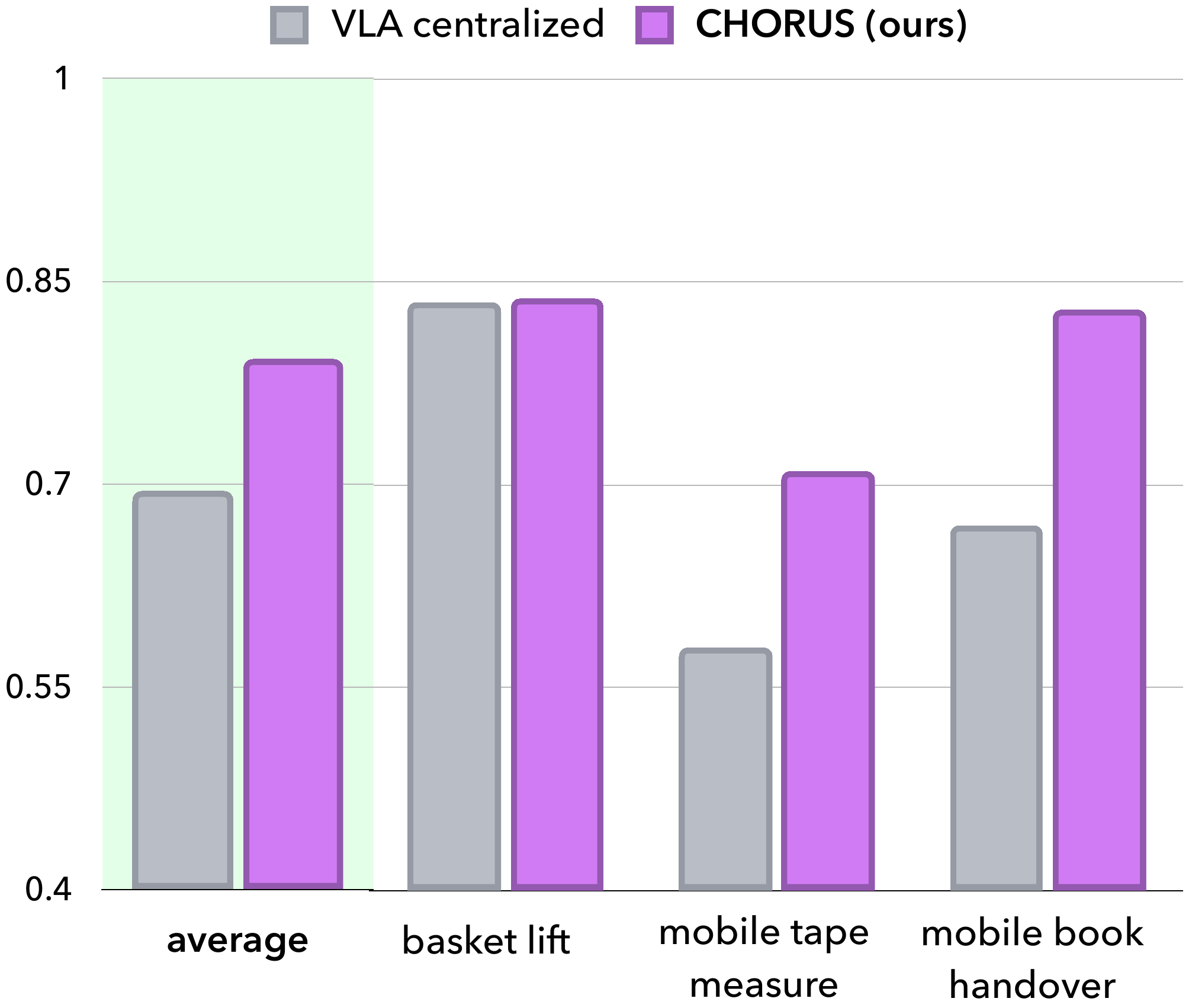}
  \caption{\textbf{Comparison to centralized policy.} Overall, CHORUS outperforms the centralized policy in mean success rate, despite the latter conditioning on all robots' observations.}
  \label{fig:centralized-comparison}
  \vspace{-1em}
\end{wrapfigure}
We train a centralized baseline as a single $\pi_{0.5}$ policy conditioned on the combined observation space of both robots. This requires a shared control rate across the team; see Appendix C for an analysis of our resampling approach. Across the three tasks, CHORUS matches or exceeds this baseline (Figure~\ref{fig:centralized-comparison}), despite conditioning only on a robot's local observations at inference.  We attribute this gap to two main sources: \textbf{distribution shift from pretraining and increased input dimensionality.} Adapting $\pi_{0.5}$ for centralized collaboration requires us to push it further away from its pretraining distribution, precluding it from achieving this upper bound in practice. $\pi_{0.5}$ is pretrained with four camera slots tied to a single mobile manipulator's front, back, base, and wrist views. Centralized coordination instead requires populating these slots with two top views and two wrist views of \emph{different robots}, breaking the semantic correspondence learned during pretraining. The centralized setup also requires conditioning on the team's entire state, so any slight desynchronization between teammates pushes the input further from the data distribution. These shifts are consistent with prior findings~\cite{dehaanCausalConfusionImitation2019} that behavior cloning performance can degrade as the input dimension grows. Since CHORUS conditions on only one robot's observations at a time, its inputs more closely reflect pretraining, allowing it to better retain its priors compared to the centralized baseline.

\subsection{Scaling to three-robot teams (Q4)}

Finally, we evaluate whether our framework extends to teams larger than two without any architectural change. We train a single CHORUS policy for a three-robot team of Kinova and YAM mobile manipulators. While the Kinova opens the door, the left YAM (YAM-1) and right YAM (YAM-2) must lift a basket, wait for the Kinova to clear the door, then navigate through the doorway to deliver the clothes in the bedroom (Figure~\ref{fig:tasks}). The task is hard for two reasons. First, \textit{YAM-1 has no rear-facing camera and cannot see when the Kinova has cleared the doorway}; it must infer when to move from YAM-2's behavior alone. Second, as YAM-2 navigates the doorway, it must adjust to avoid colliding with the door-frame, requiring YAM-1 to pause to prevent the basket from slipping. \textbf{In this three-robot setting, CHORUS attains a 90\% success rate}, demonstrating how our training recipe scales to more robots without modification.

%% file: sections/conclusion.tex
With CHORUS, we provide a recipe for adapting a VLA policy for decentralized multi-robot teams. We show that a pretrained backbone meaningfully improves collaborative manipulation, and that sharing a set of weights across robots offers reactivity benefits along with training efficiency. Lastly, we demonstrate that our framework scales gracefully to larger and heterogeneous teams.

\paragraph{Limitations.} While CHORUS enables multi-robot collaboration on many tasks, we first acknowledge that some real-world tasks do require strict instantaneous synchronization, i.e. opening two grippers at the exact same control step. These tasks necessitate centralized control and lie outside the scope of methods that can act on local observations. Second, because each robot acts on its own observations, demonstrations must contain enough information for the robot to successfully complete the task in a decentralized manner. Finally, existing large-scale manipulation datasets are overwhelmingly single-robot~\citep{oneillOpenXEmbodimentRobotic2024, khazatskyDROIDLargeScaleInTheWild2024, walkeBridgeDataV2Dataset2023}; a shared community effort to collect collaborative data would unlock significant progress in scaling multi-embodiment collaboration.

%% file: sections/appendix.tex
\appendix
 More information and videos can be found on our website: \href{https://chorus-model.github.io}{chorus-model.github.io}. 
\section{Training Details}
\label{app:training}
 
We finetune the $\pi_{0.5}$ policy with LoRA adapters on both the vision-language backbone and the action expert. Table~\ref{tab:training-hparams} lists the full set of hyperparameters. All policies are trained on a single NVIDIA H100 GPU.
 
\begin{table}[h]
\centering
\caption{Training hyperparameters.}
\label{tab:training-hparams}
\begin{tabular}{ll}
\toprule
Hyperparameter & Value \\
\midrule
Optimizer & AdamW \\
Batch size & 64 \\
Training steps & 20{,}000 \\
Training chunk size & 50 \\
Peak learning rate & $2.5\times10^{-5}$ \\
Final learning rate & $2.5\times10^{-6}$ \\
Decay steps & 30{,}000 \\
Warmup steps & 1{,}000 \\
LoRA rank (VLM) & 16 \\
LoRA rank (action expert) & 32 \\
Compute & $1\times$ NVIDIA H100 \\
\bottomrule
\end{tabular}
\end{table}
 
 
\section{Evaluation Details}
\label{app:eval}
 
\begin{figure}[t]
\centering

\begin{tabular}{cccc}
\includegraphics[width=0.22\linewidth]{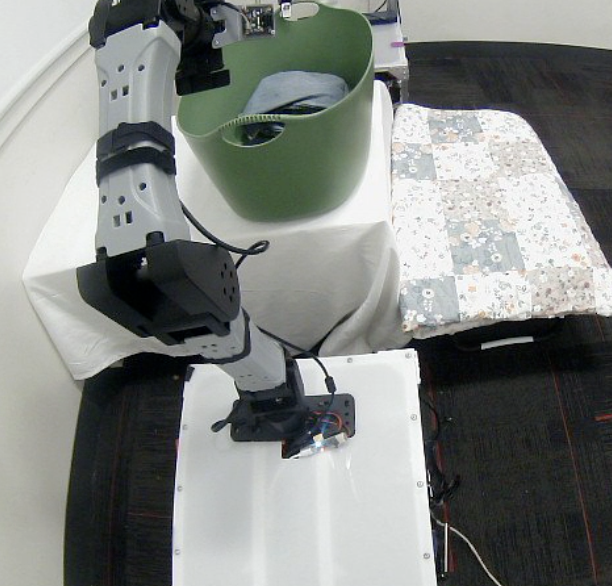} &
\includegraphics[width=0.22\linewidth]{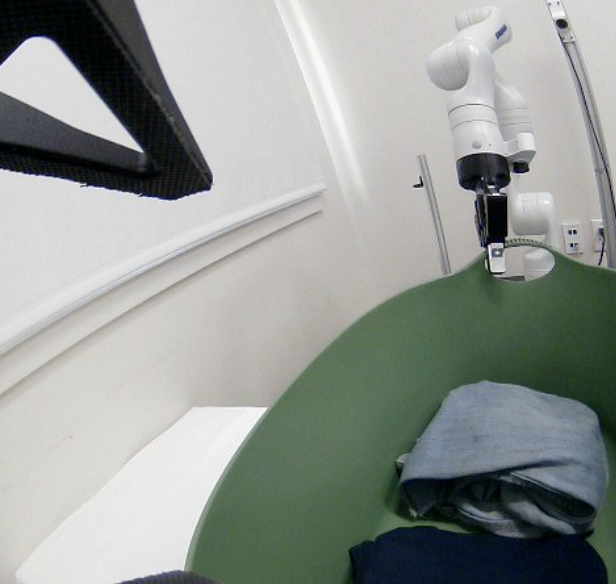} &
\includegraphics[width=0.22\linewidth]{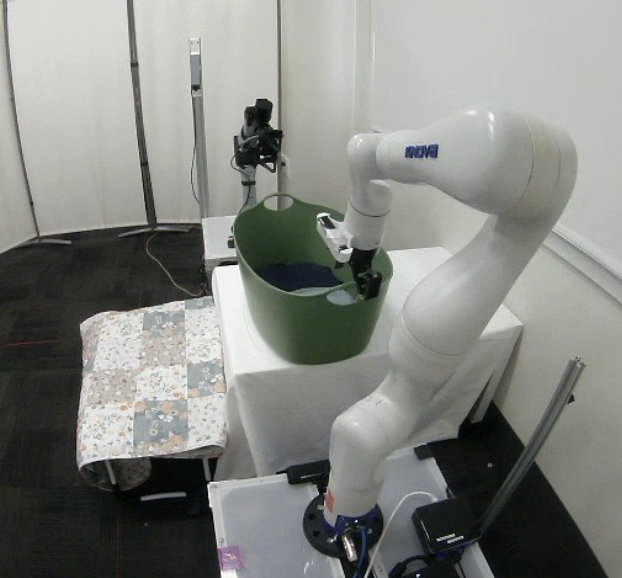} &
\includegraphics[width=0.22\linewidth]{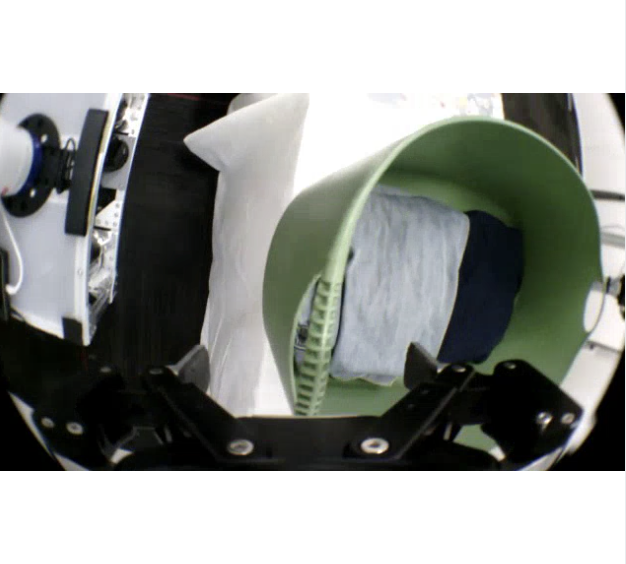} \\
\includegraphics[width=0.22\linewidth]{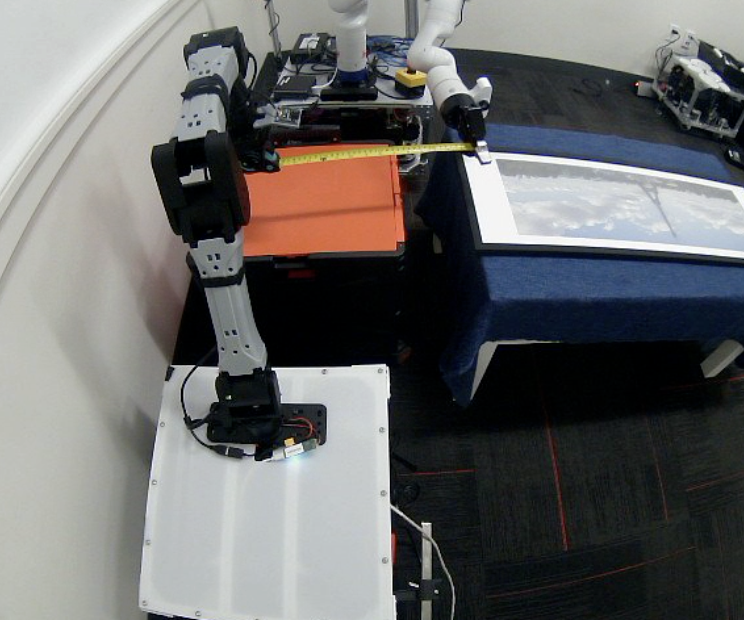} &
\includegraphics[width=0.22\linewidth]{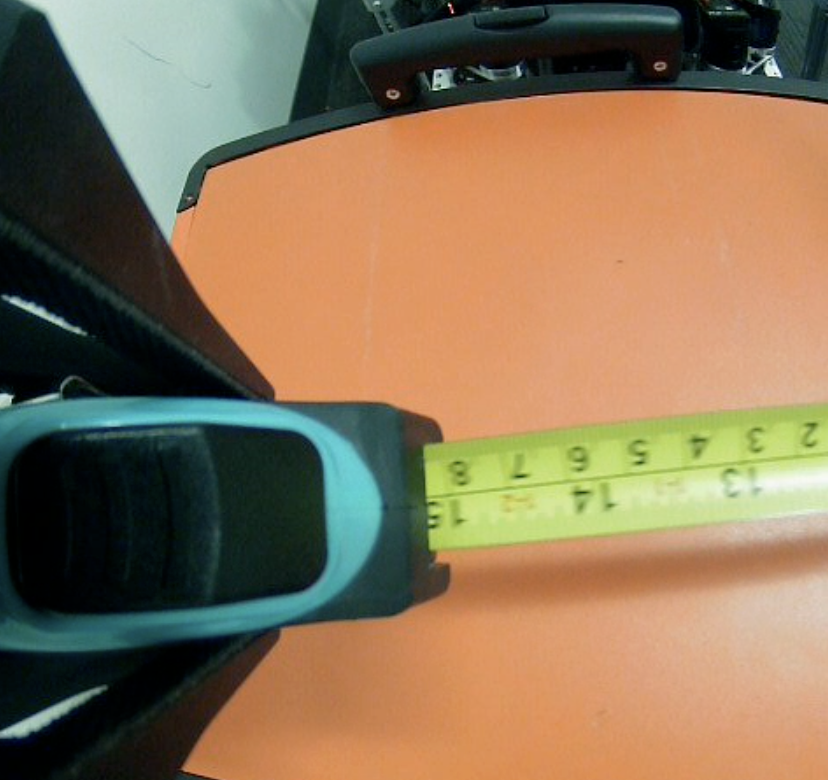} &
\includegraphics[width=0.22\linewidth]{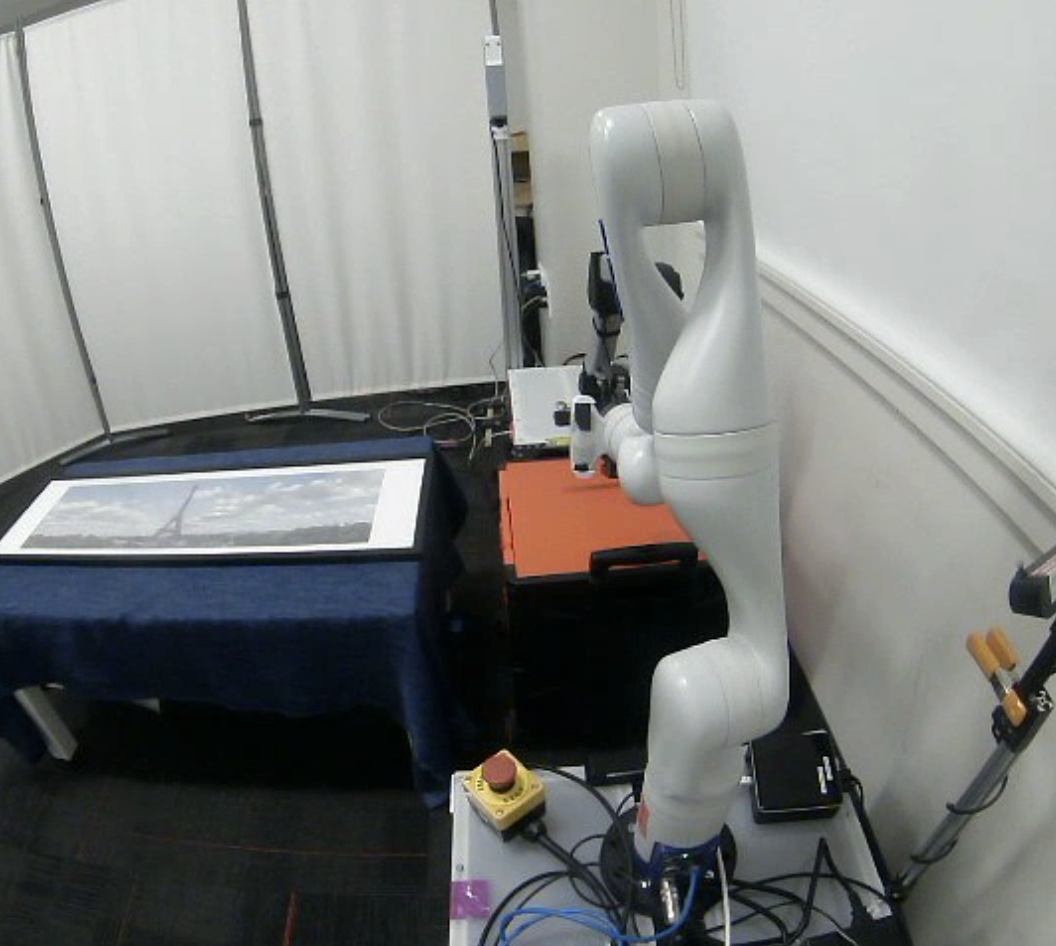} &
\includegraphics[width=0.22\linewidth]{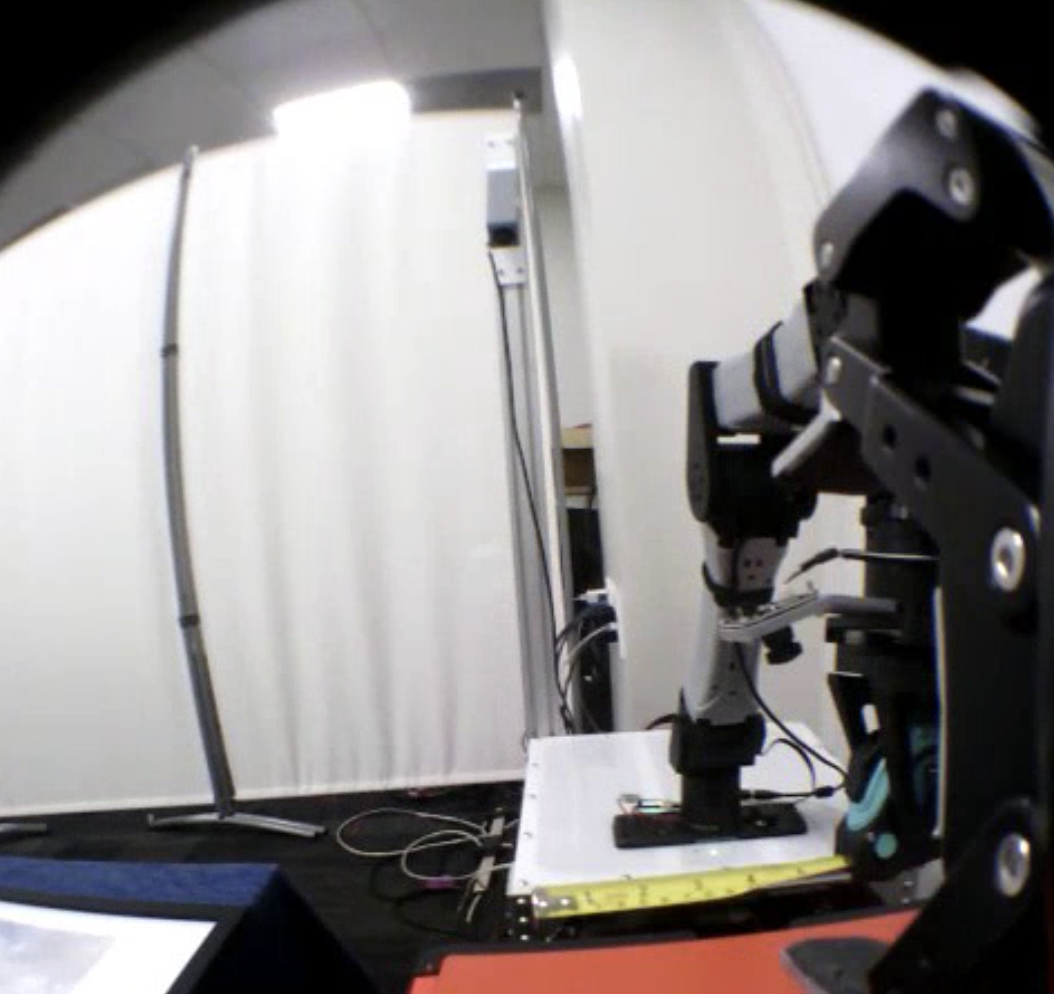} \\
\includegraphics[width=0.22\linewidth]{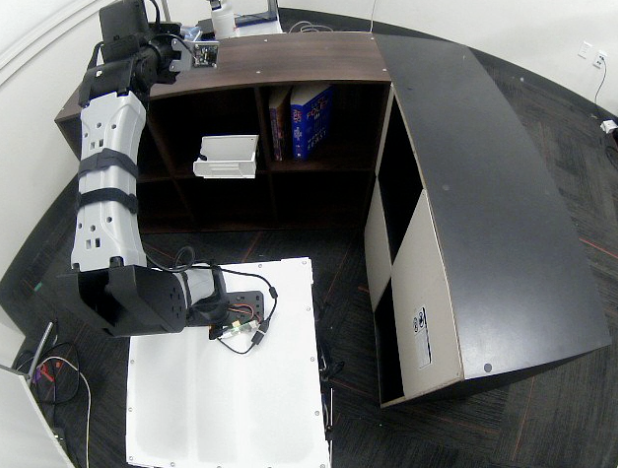} &
\includegraphics[width=0.22\linewidth]{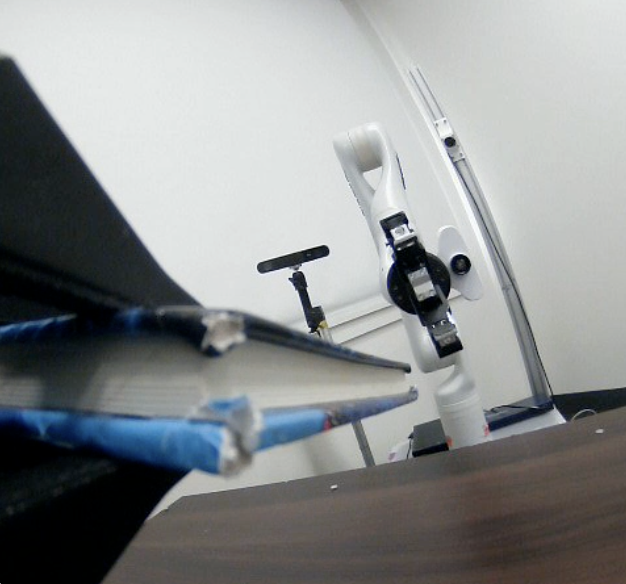} &
\includegraphics[width=0.22\linewidth]{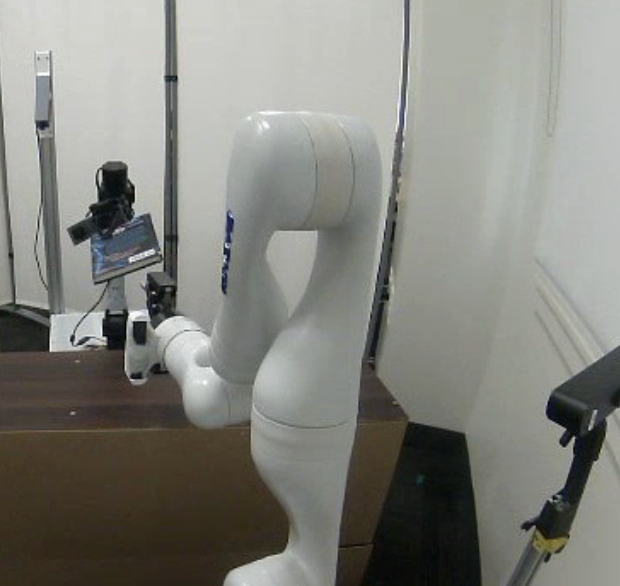} &
\includegraphics[width=0.22\linewidth]{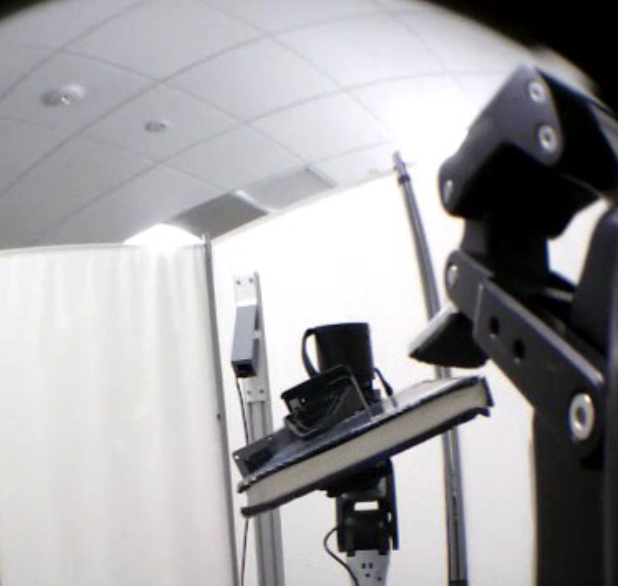} \\
\end{tabular}
\caption{Egocentric observations across tasks. Each row corresponds to the basket lift, tape measure, and book handover task respectively. The camera view order from left to right is the ARX/YAM top view, ARX/YAM wrist view, Kinova top view, Kinova wrist view. 
}
\label{fig:robot-views}
\end{figure}

\paragraph{Robot platforms and control rates.}
The basket task uses a 7-DoF ARX arm and an 8-DoF Kinova manipulator, both operating at 10~Hz. The tape measure and handover tasks use a 7-DoF YAM mobile manipulator and an 11-DoF Kinova mobile manipulator (arm plus base). The YAM operates at 30~Hz and the Kinova at 15~Hz. For the three-robot task, we add another YAM mobile manipulator, though in this case, both YAMs are 10-DoF, since we use the bases of the YAM robots. For the two-robot tasks, we include images from each robot's camera view in Figure~\ref{fig:robot-views}.

\paragraph{Deployment compute and action chunking.}
We host all policies on an NVIDIA RTX 5090 GPU. For the basket task, each robot executes a synchronized action chunk of size 30. For the remaining tasks, the YAM(s) execute chunks of 40 and the Kinova chunks of 20 to account for their control frequencies that differ by the same factor of 2.
 
\paragraph{Basket task.}
We collect 43 demonstrations. The basket position varies within two to three inches across the table. For evaluation, we run ten rollouts per policy. Four of the ten include distractors (clothing placed in the basket) that are absent from the training data. Success requires both robots to grasp their end of the basket and place it fully on the mat; a rollout receives half credit when only one robot secures its end.
 
\paragraph{Tape measure task.}
We collect 29 demonstrations. The tape measure position varies across the desk, and the picture frame varies within roughly one inch. We run twelve rollouts, half of which contain distractors. Success requires the Kinova to take the tape from the YAM and pull it out; a rollout receives half credit when the YAM grasps the tape and the Kinova begins to reach.
 
\paragraph{Handover task.}
We collect 45 demonstrations. The book position varies across three positions during data collection, and the Kinova varies across two positions. We run eighteen rollouts, six of which contain a distractor (extra book placed on the cabinet). Success requires the Kinova to grasp the book from the YAM and the YAM to release it; a rollout receives half credit when the YAM holds the book and the Kinova reaches but fails to grasp.
 
\paragraph{Three-robot transport task.}
We collect 34 demonstrations. The two YAMs train on all 34. The Kinova trains on only 18, as a portion of the door-opening data was corrupted. Note that this asymmetry is another reason why decentralized training is more desirable: a centralized policy would have to discard the extra YAM demonstrations to match the Kinova at 18 demos. The Kinova and both YAM robots start from the same initial position, while the initial basket position varies within a couple inches. We run ten rollouts. Success requires the Kinova to open the door and retreat while the two YAMs grasp the basket and move it through the doorway into the room, without collisions.
 
\section{Centralized Baseline Details}
\label{app:centralized}
 
A centralized VLA baseline conditions on a team-wide observation and produces a combined action for the team in a single forward pass, which requires all robots to run at a common control frequency. The YAM runs at 30~Hz and the Kinova at 15~Hz. We evaluate both ways of reconciling this: upsampling the Kinova to 30~Hz and downsampling the YAM to 15~Hz. We evaluated both methods for the handover task, and  found that both methods perform similarly, (Table~\ref{tab:centralized}), so we adopt the downsampling variant for the tape measure comparison in Figure~4.
 
\begin{table}[h]
\centering
\caption{Centralized frequency reconciliation on the handover task.} 
\label{tab:centralized}
\begin{tabular}{lc}
\toprule
Method & Success \\
\midrule
CHORUS & 0.833 \\
VLA Centralized (downsample) & 0.611 \\
VLA Centralized (upsample) & 0.611 \\
\bottomrule
\end{tabular}
\end{table}
 
\section{Baseline Justification}
\label{app:baseline-justification}
 
We compare against a decentralized diffusion policy baseline to isolate the contribution of the VLA backbone. The alternatives below each relax decentralization or specialize to a specific task, which makes them unsuitable as controlled baselines in our setting.
 
\textbf{MIMIC-D}~\cite{dongMIMICDMultimodalImitation2025} trains decentralized diffusion policies for multi-robot coordination. Two assumptions separate it from our setting. It conditions on proprioceptive state at evaluation, which relaxes the requirement of full decentralization, and it is demonstrated without image conditioning. Its results are primarily in simulation, with a single real-world result that fixes the object position. Reproducing it here would require conditioning each policy on the images of both robots, which breaks the decentralization assumption we maintain throughout.
 
\textbf{Latent Theory of Mind}~\cite{heLatentTheoryMind2025a} also pairs two separate diffusion policies. It also maintains two assumptions. It relies on a shared third-person camera view across the robots, requiring inter-robot communication at inference, and it runs an online alignment procedure that iteratively updates the robots' actions through a sheaf Laplacian. CHORUS uses neither iterative updates nor communication, operating under full decentralization. Our diffusion baseline can be viewed as this method's core architecture with its inference-time assumptions removed.
 
\textbf{GCo}~\cite{shaoulCollaborativeMultiRobotNonPrehensile2025} and related work coordinate multi-robot manipulation by pairing learned interaction primitives with classical multi-robot planning. These methods decompose the problem, routing objects and allocating robots through a centralized planner with access to global world state, and learn only short-horizon contact and pushing strategies. CHORUS removes the central coordinator entirely. Each robot conditions on its own local observations and produces its own actions, with coordination arising from training data rather than an explicit assignment and planning stage. This difference also makes such methods ill-suited as a baseline in our setting. They presuppose a global state estimate and a homogeneous robot team, whereas we target heterogeneous embodiments acting under partial, local observations. They are further evaluated on planar, non-prehensile pushing in simulation, a task class that does not exercise the contact-rich, real-world coordination our experiments demand. Adapting them to our problem would require replacing their core planning machinery, leaving little of the original method intact to compare against.
 
\textbf{decPLM}~\cite{panditMultiQuadrupedCooperativeObject2025} trains an MLP policy conditioned on state alone through a phased, centralized procedure. Its central contribution is a reward design built around target contact points on the box for each robot, with additional terms penalizing box acceleration, action roughness, and large torques and joint motions. CHORUS differs in three ways. Our training is fully decentralized. Our conditioning is simpler, requiring neither contact information nor a structured reward. And their method is specific to box pickup and transport, so it does not extend to the range of tasks we evaluate.
 
The closest remaining analog is learning-based bimanual manipulation \cite{fuMobileALOHALearning2024}, which likewise presumes a shared third-person view and a single policy commanding both arms. This is the structure of our centralized baseline, which conditions on a team-wide observation and produces a joint action for the whole team. These assumptions, of course, are aside from the decentralized setting we study, in which each robot observes and acts only on its own stream. Decentralized diffusion is the one baseline that holds the decentralization assumption fixed while removing the VLA backbone, which is what makes it the controlled comparison for our central claim.